\documentclass[runningheads]{llncs}
\usepackage[T1]{fontenc}
\usepackage{indentfirst}
\usepackage[colorlinks=true, urlcolor=blue, linkcolor=red]{hyperref}
\usepackage{graphicx}
\usepackage{amsmath}
\usepackage[section]{placeins}
\usepackage{booktabs} 
\begin{document}

\title{Self-supervised Fusarium Head Blight Detection with Hyperspectral Image and Feature Mining}

%
%
\author{Yu-Fan Lin  \and Ching-Heng Cheng  \and Bo-Cheng Qiu  \and Cheng-Jun Kang  \and \\Chia-Ming Lee  \and Chih-Chung Hsu  \thanks{means the corresponding author.}}
\authorrunning{Y.F. Lin et al.}
%
\institute{Institute of Data Science, National Cheng Kung University, Taiwan\\
\email{\{aas12as12as12tw, henry918888, a36492183, cjkang0601, zuw408421476\}@gmail.com}\\
\email{\inst{*}cchsu@gs.ncku.edu.tw}}
\maketitle              
\begin{abstract}
Fusarium Head Blight (FHB) is a serious fungal disease affecting wheat (including durum), barley, oats, other small cereal grains, and corn. Effective monitoring and accurate detection of FHB are crucial to ensuring stable and reliable food security. Traditionally, trained agronomists and surveyors perform manual identification, a method that is labor-intensive, impractical, and challenging to scale. With the advancement of deep learning and Hyper-spectral Imaging (HSI) and Remote Sensing (RS) technologies, employing deep learning, particularly Convolutional Neural Networks (CNNs), has emerged as a promising solution. Notably, wheat infected with serious FHB may exhibit significant differences on the spectral compared to mild FHB one, which is particularly advantageous for hyperspectral image-based methods. In this study, we propose a self-unsupervised classification method based on HSI endmember extraction strategy and top-K bands selection, designed to analyze material signatures in HSIs to derive discriminative feature representations. This approach does not require expensive device or complicate algorithm design, making it more suitable for practical uses. Our method has been effectively validated in the Beyond Visible Spectrum: AI for Agriculture Challenge 2024. The source code is easy to reproduce and available at \hyperlink{https://github.com/VanLinLin/Automated-Crop-Disease-Diagnosis-from-Hyperspectral-Imagery-3rd}{https://github.com/VanLinLin/Automated-Crop-Disease-Diagnosis-from-Hyperspectral-Imagery-3rd}.

\end{abstract}

\section{Introduction}

With the continuous increase in the global population, food supply is a critical issue \cite{c1,c2,c3}, especially for developing countries. Every year, a significant amount of crops is lost worldwide due to pests and diseases \cite{Jasonnatureplants2015}. Accurate and timely detection of crop diseases is crucial for global agricultural production.

Effective diagnosis and control measures rely on symptom identification and severity assessment. Traditionally, these tasks are achieved through observation and estimation by humans, which may be impractical for large-scale monitoring due to high costs. Some prior-based approaches commonly used handcrafted features to identify features on land, such as the Normalized Difference Vegetation Index (NDVI) \cite{rouse1974ndvi} and the Normalized Green-Red Difference Index (NGRDI) \cite{tucker1979red}, are calculated based on the reflectance composition of specific spectral bands. 

However, even though these indices may have the potential to distinguish features to some extent, they may fail in certain circumstances because they rely on a small number of bands within RS images or multispectral images (MSIs). Relying on such a limited number of bands does not always yield satisfactory results because spatial spectrum correlation and information are not fully exploited.

Fortunately, recent advancements in HSI imagery have made it possible to develop automated image-based crop disease diagnostic methods. Endmember extraction is a crucial process in HSI and RS that involves identifying pure spectral signatures from mixed pixel data \cite{end1,end2}. This process is essential for tasks such as material identification, classification, and abundance estimation in various applications like environmental monitoring, agriculture, and mineral exploration \cite{c2}. With the development of HSI analysis, several methods designed for material recognition are introduced, such as endmember extraction, HSI unmixing and non-negative matrix factorization. These methods are mainly optimization-based, effectively integrating the low-rank prior and sparsity prior of HSI.

Moreover, with the development of deep learning paradigms represented by CNNs and Vision Transformers \cite{VIT}, using deep neural networks for detection is a promising direction \cite{official}. Devadas et al. \cite{DevadasPrecisionAgric2009} used 10 widely-used vegetation indices (VIs) to identify different types of rust in wheat leaves, and the results showed that some indices are effective. Shi et al. \cite{ShiPrecisionAgric2009} applied wavelet techniques based on HSI to detect yellow rust on wheat leaves \cite{FarrellImpact2005}. The learning-based methods are gradually dominating the HSI analysis for crop disease recognition. As our experiments and analysis shown, the spectral of diseased FHB images are differ from mild one. In other words, the endmember extraction would be effective to boost our model performance and robustness due to the extracted features may be discriminate. 

In this report, we propose a self-supervised method for FHB detection. First, we analyze the intensity of each band in HSIs and reduce the complexity. Secondly, top-K bands selection with the guidance of pseudo-label generated by the K-means clustering is used to extract the discriminative features. Afterwards, the arbitrary classifier is assigned for FHB detection due to these effective key features are obtained. Compared to some detection methods based on CNNs, our proposed method is more suitable for large-scale general applications because it does not require the use of GPUs or the collection of a large number of expensive HSIs. We believe this study will provide valuable insights for future FHB detection.


\section{Methodolodgy}

\begin{figure}[t] 
	\begin{center}
 		\includegraphics[width=0.9\linewidth]{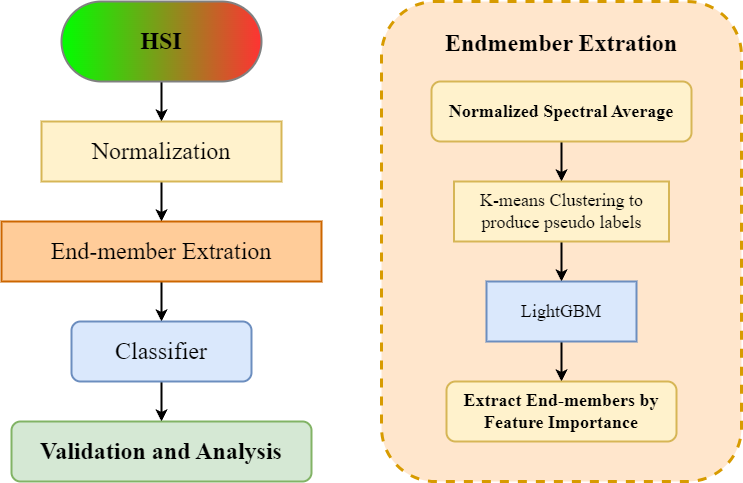} 
	\end{center}
        \vspace{-0.15in} 
	\caption{The overall pipeline of the proposed method.}
	\label{fig:pipeline}
\end{figure}

In this section, we introduce a strategy to effectively utilize the spectral information of image-level HSIs using the simple endmember extraction strategy with top-K bands selection. Afterward, the simple classifier is employed to detect the given HSI is suffered from mild-FHB or serious-FHB, as shown in Figure \ref{fig:pipeline}.

\subsection{Top-k bands selection for Endmember Extraction} Given the unique characteristics of HSIs, where different materials exhibit distinct reflectance values at the same bands, we propose an efficient approach for FHB detection that bypasses the need for complex HSI unmixing techniques traditionally used in endmember extraction.

Before the endmember extraction, the normalization and spectral averaging are used for reducing data complexity and remove the noises and redundancy within HSI. Because of the FHB detection can be regarded as the type of coarse-level HSI classification, there is no need for keeping fine-grained feature well. Therefore, normalization strategy is beneficial to the better performance and robustness for FHB detection. 

As we mentioned in the previous section, the traditional metrics for diseased crops recognition, such as NDVI and NGRDI, are insufficient to handle more diverse scenes and the complexity of hyperspectral imagery (HSI). To reduce the complexity of HSI and streamline the recognition procedure, we introduce a top-K band selection method for key bands extraction. We begin by utilizing K-means clustering to generate pseudo-labels, providing valuable guidance. The normalization and spectral averaging techniques enable K-means clustering to achieve excellent clustering results.

\begin{figure}[t] 
	\begin{center}
 		\includegraphics[width=0.95\linewidth]{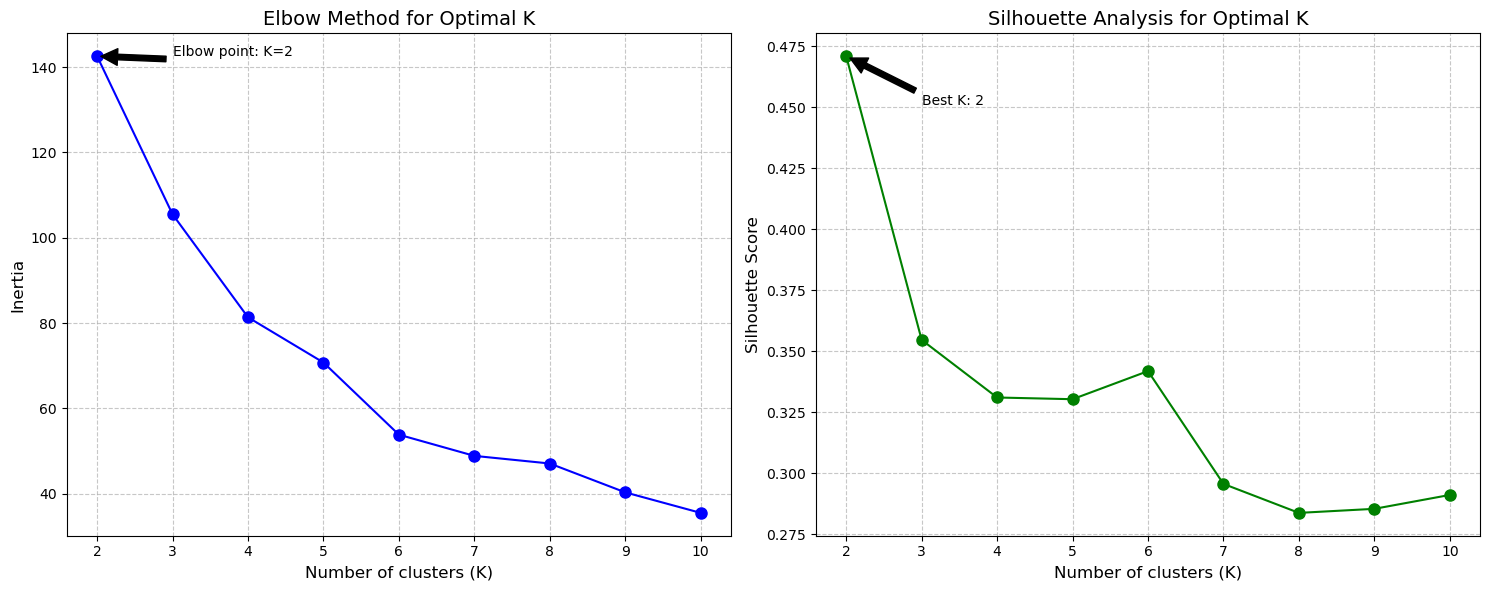} 
	\end{center}
        \vspace{-0.15in} 
	\caption{Elbow method (left) and Silhouette Analysis (right). Both methods show that when K is set to 2 as the best selection.}
	\label{fig:best_k}
\end{figure}

To determine the optimal number of clusters (K), we employed two complementary methods: the Elbow Method and Silhouette Analysis, as illustrated in Figure \ref{fig:best_k}. The Elbow Method, shown in the left graph, plots the inertia (within-cluster sum of squares) against the number of clusters. The optimal K is identified at the "elbow" point where the rate of decrease in inertia begins to level off, occurring at the optimal K is selected in our analysis. Correspondingly, the Silhouette Analysis, depicted in the right graph, measures how similar an object is to its own cluster compared to other clusters. The highest Silhouette score indicates the same result for the best clustering. The consistency between these two methods reinforces our confidence in selecting the optimal number of clustered groups as the best value.

Then, we utilized K-means clustering to generate pseudo-labels, providing guidance for our analysis. Due to normalization and spectral averaging techniques, K-means clustering achieved excellent clustering results. These pseudo-labels were subsequently used for feature importance mining. Based on this analysis, we selected the top-K important bands to serve as endmembers for mild or serious FHB recognition. Unlike traditional endmember extraction methods in HSI analysis, which primarily rely on optimization-based approaches with dense mathematical computations, our method is simpler. Its simplicity stems from the direct extraction of discriminative features, streamlining the process while maintaining effectiveness.


\subsection{Simple Classifier for Fusarium Head Blight Detection} Because the robust and compact features are yielded, the classifier we used can be arbitrary. We use LightGBM \cite{KelightGBM2017} as our primary classifier because LightGBM excels in handling high-dimensional data with efficiency and robustness. It is especially beneficial in detecting FHB infections using HSI, which is usually a high-dimensional cube data, helping to mitigate the need for chemical treatments for treating wheat in the future.  For the comprehensive analysis, we also used Support Vector Machine (SVM) in our experiment, it will be discussed in the next session.

\section{Experiment Results}

\indent\textbf{Data Description.} The dataset we used is from \cite{official}. These HSIs were acquired using a DJI M600 Pro UAV system equipped with an S185 snapshot hyperspectral sensor, capturing reflectance from 450-950nm with a spectral resolution of 4nm. The raw data included a $1000 \times 1000$ px panchromatic image and a $50 \times 50$ px hyperspectral image with 125 bands. Due to noise interference, the first 10 and last 14 bands were excluded, leaving 101 bands. All images were captured at a 60 meter altitude, providing a spatial resolution of approximately 4cm per pixel. While in the training stage, the data is divided into blocks with a size of $32 \times 32$ and labeled. The dataset consisted of a total of 1,696 HSIs, of which 1,006 were labeled as mild-FHB and 690 as serious-FHB. During the training phase, we randomly selected 1,611 HSIs as training data, leaving the remaining 85 for validation.

\begin{figure}[t] 
	\begin{center}
 		\includegraphics[width=0.95\linewidth]{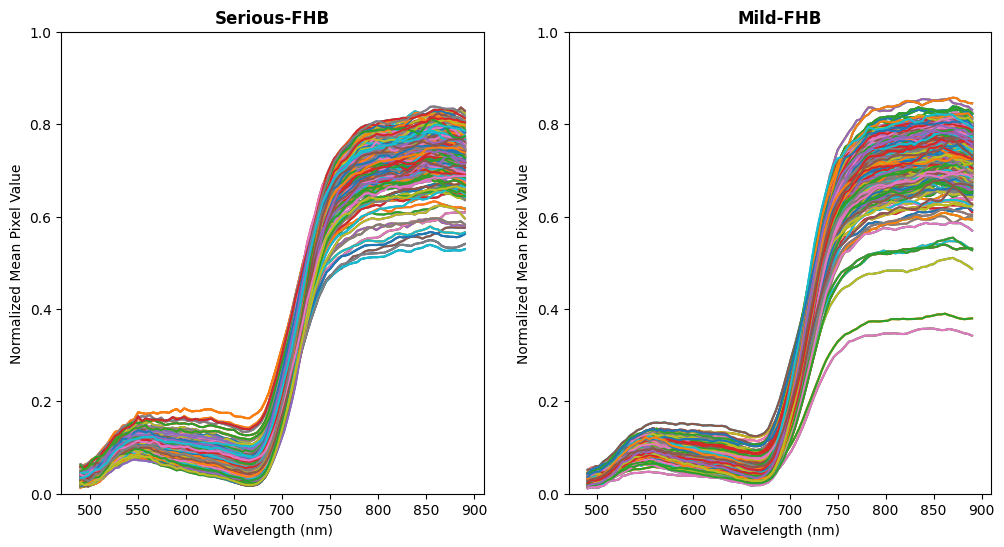} 
	\end{center}
        \vspace{-0.15in} 
	\caption{The overall normalized spectrum profiles of the FHB dataset. } 
	\label{fig:his}
\end{figure}

\begin{figure}[t] 
	\begin{center}
 		\includegraphics[width=0.95\linewidth]{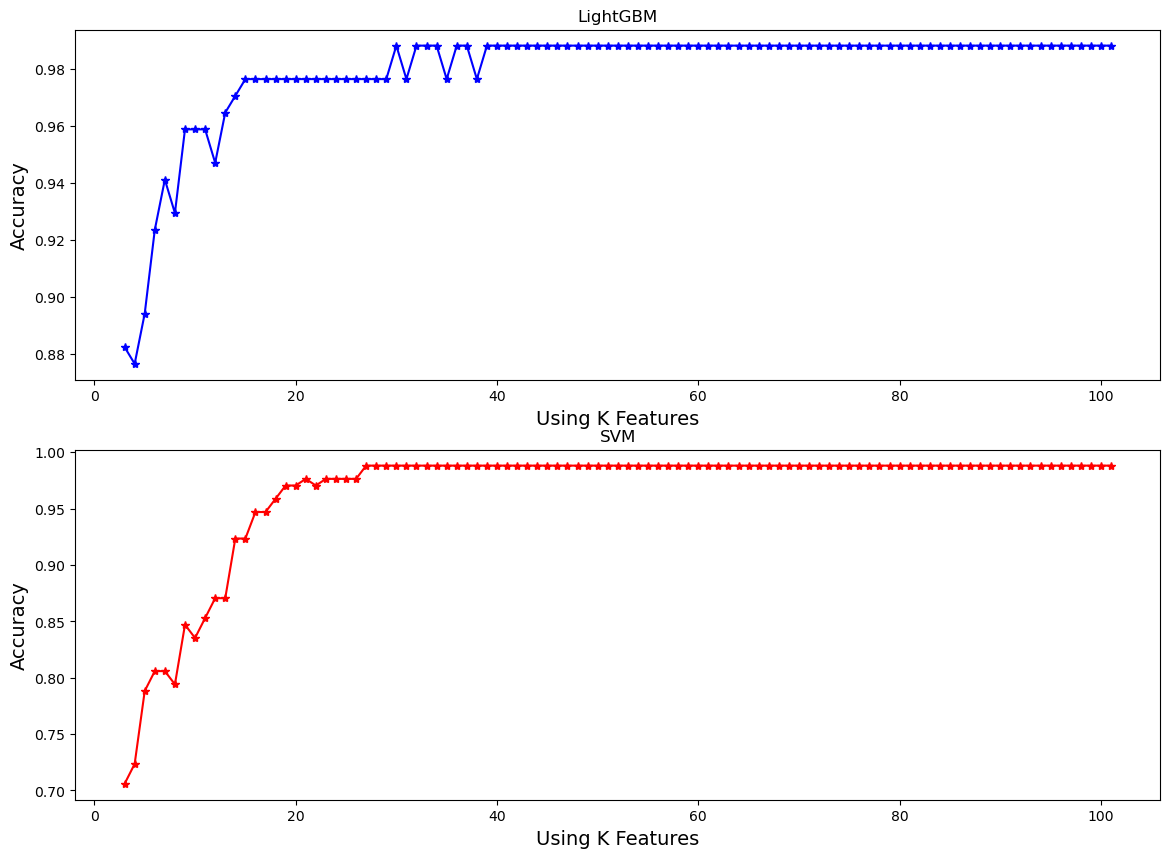} 
	\end{center}
        \vspace{-0.15in} 
	\caption{The performance comparison of top-K bands selection strategy with different classifier and selected bands.}
	\label{fig:his2}
\end{figure}

\textbf{Experiment Settings.} In the experiment, we simply used two different models, LightGBM and SVM as our classifiers, as our previously illustrated, classifier can be arbitrary due to robust and discriminative representations are extracted by the proposed self-supervised endmember extraction strategy. The hyperparameter settings are all kept default. The input HSIs are normalized for stabilizing the training and inference. The common augmentation strategies are employed, such as random rotation and random cropping. 
In this study, we use accuracy to evaluate the performance of classification for FHB detection. Accuracy is defined as the ratio of correctly classified samples to the total number of samples. Additionally, four metrics—true positives (TP), true negatives (TN), false positives (FP), and false negatives (FN)—are also can be used to calculate accuracy as follows:

\begin{equation}
\text{Accuracy} = {{\text{TP}+\text{TN}} \over {\text{TP} + \text{TN} + \text{FP} + \text{FN}}}
\end{equation}

\textbf{Top-K bands selection.} The number of selected bands is important, it may lead the different extracted feature and prediction. Our experiments revealed that setting k to approximately 30 resulted in perfect accuracy on the validation set. Figure \ref{fig:his2} illustrates that even with k is set to 3, the accuracy approaches 90\%. Notably, as the selected bands increase, the accuracy is higher. It emphasizes the critical role of selecting important bands as features for model performance.



\textbf{Evaluation and Analysis.} The performance of our method have shown in Figure \ref{fig:his2}. The proposed method have showed the outstanding performance while significantly reducing the complexity. Ultimately, the top-30 bands are used during both stages of our method, striking an optimal balance between computational efficiency and model performance.


To add the explainability of our method, we further analyzed the important bands. Figure \ref{fig:his} illustrates the spectral signatures of mild-FHB and serious-FHB crops, highlighting the important features identified by our method across the wavelength range. Interestingly, as depicted in Figure \ref{fig:his2}, our analysis reveals that these critical bands are not uniformly distributed but rather cluster within specific wavelength ranges. Notably, we observed significant concentrations in the 700-750 nm and 800-875 nm ranges, corresponding to the green and red spectra, respectively. This clustering of important features suggests that a large portion of the hyperspectral data may be redundant for FHB detection. Instead, effectively leveraging information from these two key spectral regions appears to be crucial for accurately identifying FHB.

These finding not only validates the effectiveness of our feature selection approach but also offers valuable insights into the spectral characteristics most relevant for FHB detection. The clear difference between mild-diseased and serious-diseased spectral patterns, coupled with the top-K bands selection for endmember extraction, demonstrates the robustness and discriminative power of our method.

\section{Conclusion}
In this paper, we proposed a self-supervised method for FHB detection based on hyperspectral imagery. The core of our method lies in  endmember extraction, followed by a novel top-K bands selection. With the selected significant bands, we effectively reduced the dimensionality of given HSI without compromise between complexity and performance for FHB detection. Because of the above methods are self-supervised and without the needs for the expensive device for storage and deep neural networks, we believe our methods are more easy and feasible to the practical uses and applications. The efficacy of the proposed method has been demonstrated in our experiments and the Beyond Visible Spectrum: AI for Agriculture Challenge 2024. 



\begin{figure}[htbp]
	\begin{center}
 		\includegraphics[width=0.95\linewidth]{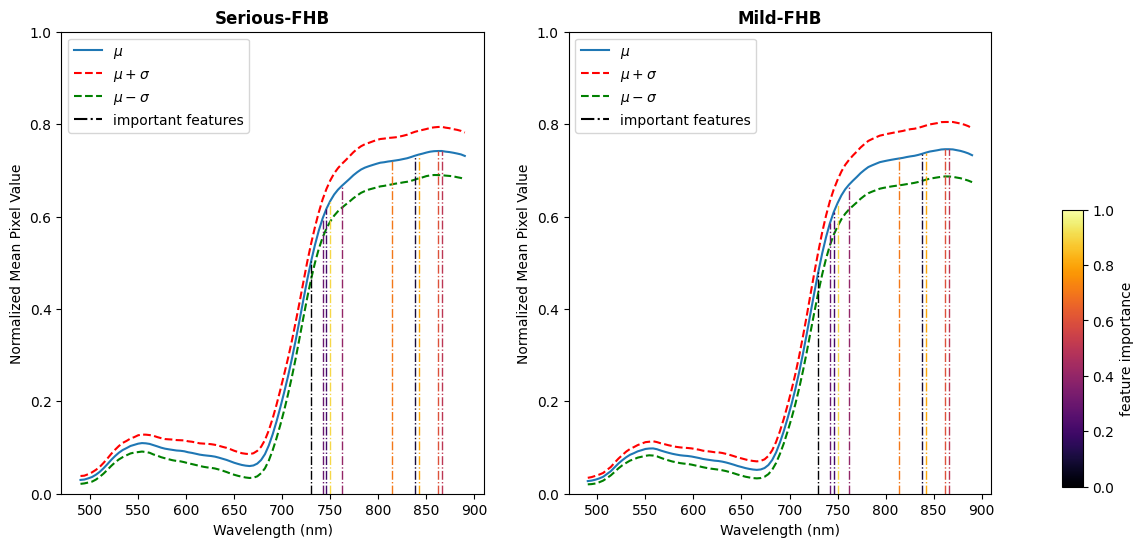} 
	\end{center}
        \vspace{-0.15in} 
	\caption{The indication of important bands on both serious-FHB and mild-FHB spectrum profiles. } 
	\label{fig:his3}
\end{figure}



\end{document}